# Times Series Forecasting for Urban Building Energy Consumption Based on Graph Convolutional Network

Yuqing Hu [a], Xiaoyuan Cheng [a], Suhang Wang [b], Jianli Chen [c,] *, Tianxiang Zhao [b], Enyan Dai [b]

[a] Department of Architectural Engineering, The Pennsylvania State University, University Park, PA 16802, USA
[b] College of Information Science and Technology, The Pennsylvania State University, University Park, PA 16802, USA
[c] Department of Civil and Environmental Engineering, University of Utah, Salt Lake City, UT 84112, USA

## ABSTRACT

The world is increasingly urbanizing and building industry accounts for more than 40% of energy consumption in the United States. To improve urban sustainability, many cities adopt ambitious energy-saving strategies through retrofitting existing buildings and constructing new communities. In this situation, an accurate urban building energy model (UBEM) is the foundation to support the design of energy-efficient communities. However, current UBEM are limited in their abilities to capture the inter-building interdependency due to its dynamic and non-linear characteristics. Those models either ignored or oversimplified these building interdependencies, which can substantially affect the accuracy of urban energy modeling. To fill the research gap, this study proposes a novel data-driven UBEM synthesizing the solar-based building interdependency and spatial-temporal graph convolutional network (ST-GCN) algorithm. Especially, we took a university campus located in downtown of Atlanta as an example to predict the hourly energy consumption. Furthermore, we tested the feasibility of proposed model by comparing the performance of the ST-GCN model with other common time-series machine learning models. The results indicate that the ST-GCN model overall outperforms than all others. In addition, the physical knowledge embedded in the model is well interpreted. After discussion, it is found that data-driven models integrated engineering or physical knowledge can significantly improve the urban building energy simulation.

## INTRODUCTION

The world is rapidly urbanizing. It is reported that urban areas will account for around 70% of energy and $CO_2$ emissions and house 66% of the world's population by 2050. Urban sustainability is a defining challenge in the 21st Century. Many regions have adopted ambitious sustainable goals to reduce emissions and energy usage. For example, New York sets 45% total emission reduction from 1990 levels by 2030 [1]; the California government has established the Clean Energy and Pollution Reduction Act to reduce the green gas emission by 40% from 1990 levels, and up to 80% by 2050 [2]. Since building industry accounts for around 41% of energy and 76% electricity in the United States [3]. Accurately forecasting urban building energy consumption is critical to accelerate the transformation of sustainable cities. However, individual building energy consumption not only depends on its own properties and characteristics, but also influenced by surrounding buildings, especially in high-density urban areas [4-7]. Those studies have demonstrated the significance of building interdependency on building energy consumption. These interdependencies can refer to shadowing impacts, short/long-wave radiation impacts, wind flow impact, and human mobility between buildings. Previous studies have indicated that inter-building impacts can increase the temperature of certain exterior surfaces up to 10℃ as well as 3%, 15%, 40% of annual, daily and summer peak cooling demand, respectively [5, 8]. Therefore, modeling the interdependency among buildings is critical for developing representative predictive building energy models in urban context. Capturing the interdependency is challenging due to its non-linear, high-dimension and dynamic characteristics [7]. Those factors adversely influence the simulation accuracy and computational overhead and limit models scalability. In this situation, most interdependencies in previous urban building energy models (UBEM) are oversimplified [9] or completely neglected [10] and further studies need to be conducted.

Generally, UBEM can be grouped into categories: physics-based and data-driven modeling [8]. The accuracy of physics-based is not reliable due to the assumed building properties and environmental parameters. The broadly agreed variance of the physics-based model is about 30% without model calibration [11], which limits the model application merely to evaluate relative energy efficiency rather than predict absolute energy consumption [12]. Even though a well-calibrated physics-based model is robust in forecasting building-level energy use, the time-consuming calibration process caused by the complexity of the algorithm adversely affects its scalability [4]. With the rapid development of sensor technologies and the emergence of data mining techniques, data-driven UBEMs are increasingly drawing attention [13, 14]. Many machine learning algorithms, e.g., linear regression [15], support vector machine [15, 16], and neural network [7, 15], have been used to predict building energy consumption. However, most of these data-driven models focus on individual buildings' energy simulation and ignore interdependency among buildings [9, 10], which makes the results less reliable and explainable. In terms of the nature of data-driven models, they capture the target results based on building-level characteristics, e.g., building insulation level, thermal mass, occupant schedule. They assume that buildings are independent, which is contrary to the actual situation. Buildings interact in many aspects (direct interdependency: long-wave and short-wave radiation impact [5] [17]; indirect interdependency: wind flow and heat transfer [5]). To model the urban building interdependency, this study will demonstrate that data-driven models can employ existing knowledge to capture the interdependency in some aspect. We propose to use the graph neural network (GNN) methodology to develop a novel UBEM that include the interdependency between buildings into consideration. This approach will be compared with other machine learning models to demonstrate its superiority. The paper is structured as below: Section 2 discusses the current efforts of urban-level building energy prediction, identifies the existing knowledge gap, and indicates the potential of GNNs on filling these gaps; Section 3 identifies the methodology of this paper, including how to capture initial inter-building impacts, how to develop GNNs models, and how to assess model performance; Section 4 demonstrates the performance of GNN in the prediction of hourly building energy prediction by using a case study with 26 buildings of a university campus in the downtown area at Atlanta, Georgia and compare it with other frequently used data-driven models. Section 5 is the result discussion; followed by the conclusion and future improvement suggestions.

## LITERTURE REVIEW

To support sustainable city or community planning, many studies have put great efforts into urban building energy modeling (UBEM), which can be classified into top-down and bottom-up approaches [8, 18]. Top-down methods analyze the long-term relationships between building energy use with society socio-econometric and technological indicators [19, 20], e.g., gross domestic product, income, and technological progress [21]. which are all aggregated data without precise spatial or temporal information. By contrast, the bottom-up category quantifies the energy use of the single building and then aggregates those data to derive urban building energy use in multi-scales. This paper employs a bottom-up UBEM to discusses the possibilities of improving building energy modeling in an urban context. Bottom-up approaches can be further classified into physics-based models and data-driven models [8]. Physics-based models employ the laws of heat transfer and thermodynamics to simulate building energy consumption, which requires a detailed description of building characteristics [22, 23]. The increasing number of objects in simulation leads to booming complexity, especially considering the interactions among buildings and their local environment.

Numerous efforts have been put forth to improve physics-based UBEM in the last decades. Maintaining a good balance between reducing computational power and model accuracy has continuously been a hot spot in this field. To improve the model efficiency, many studies and tools utilized representative archetypes to describe buildings in the urban context [18], [57]. The core of those studies is to rationalize archetypes and building sampling to improve the representativeness of the selected archetypes to the overall building profiles in the study area [24, 25]. In addition, some studies employ actual building data to improve the model accuracy. For example, great contributions have made by synthesizing UBEM tools and Geographic Information System (GIS) techniques to capture geometric information of buildings [22, 24, 26, 27]. Drones [28], sensors [29], computer vision techniques [30], and other open-source datasets [31] are employed to obtain building properties. However, some building characteristics such as building shapes, thermal properties and interactions are challenging to simulate, which can significantly affect building shading, urban heat islands, and wind flow, then further directly or indirectly influence building energy consumption. Creating heat balance equations to describe those impacts and solving these equations are highly complex and time-consuming. To ensure model efficiency and scalability, existing physics-based models and tools either simplify or neglect these interactions and microclimates [9, 10]. In this situation, model accuracy cannot be guaranteed. Even though many approaches have been developed so far, model calibration is still the mainstream method for accuracy improvement. Due to the unbearable computational overhead, the calibration of physics-based UBEM is difficult to apply at a city level.

To overcome the drawbacks of physics-based models, data-driven UBEM, which benefit from adoptions of smart sensors [32], open data initiatives [33], and data mining techniques [34], have been increasingly adopted due to less demanding engineering efforts in model establishment [26]. Data-driven UBEM can be categorized into statistical methods and machine learning methods [7]. Statistical models predict building energy consumption by inputting data to an explicit mathematics function, whereas machine learning methods learn the relationships between input features and actual energy consumption data to train models [26]. As the machine learning approach requiring fewer human efforts to empower UBEM, it is becoming a hot spot in this field. Previous studies have adopted multiple machine learning algorithms in UBEM, including linear regression [35], support vector machines [36], neural networks [37], and deep learning [38]. Furthermore, the availability of public databases and platforms (e.g., DataSF[1], NYC OpenData[2], OpenStreetMap[3]) containing energy consumption and building characteristics data promotes the development of data-driven UBEM. Based on the requirement and data granularity, data-driven UBEM can run in multiple temporal resolutions [7, 39], annual, monthly, daily, or even hourly. Although many studies have argued those data-driven models are adoptable at the individual building levels, the potentials of those models can be unlocked to simulate complex building interdependency.

Nevertheless, most data-driven UBEM are insufficient in capturing building interdependency due to the complexity of the physical system. Nutkiewicz et al. [7, 38] proposed a novel network-based machine learning algorithm (ResNet) to simulate urban building energy consumption. Especially, they generated desired periodic energy data for each building from the baseline energy simulation model, which used for training the data-driven model. Then, a residual neural network was developed to simulate the 'hidden' impacts of the urban environment that are not captured in the individual building simulation. They provided a hybrid method synthesizing physical knowledge and neural network to simulate energy consumption in different spatiotemporal resolutions. However, generating the energy consumption data from a physics-based model is time-consuming and the weight of hidden layers in ResNet cannot be extracted to explain the correlations between consumption in adjacent buildings. To solve these problems, we propose a novel framework by integrating building interdependency in graph neural network (GNN) to simulate the building energy consumption on an urban scale.

Scarselli et al. [40] firstly proposed the terminology of GNN, aiming to extend existing in processing the data represented in graph domains. In a graph network, GNNs adopt the message passing mechanism, which means that the representation of each node is naturally defined by its attributes and aggregation of the adjacent nodes [41]. Due to the nature of graph network, GNN has been broadly employed in structural scenarios where data structures have explicit relationships, such as social network [42], transportation system [43], and recommendation system algorithm [44, 45]. The information propagation allows GNN models to learn building energy consumption based on its own characteristics and propagate the energy impact from neighborhood buildings via graph structure. One graph type is worth noting is the dynamic GNN [46], which has a static graph structure and dynamic input signals. To capture both kinds of information, Spatio-Temporal Graph Convolutional Network (ST-GCN) [47] first collects spatial information by GNNs, then feeds the outputs into a sequence model like sequence-to-sequence model or CNNs. Differently, STGCN [47] can collect spatio-temporal messages at the same time. They extend static graph structure with temporal connections so they can apply traditional GNNs on the extended graphs. Due to the seasonal and regional characteristics of the building energy consumption, the study takes STGCN as the main framework to model the interdependence among buildings and predict building energy use.

[1] https://datasf.org/opendata/
[2] https://opendata.cityofnewyork.us/
[3] https://www.openstreetmap.org/#map=17/35.41800/-91.98083

## METHODOLOGY

In this section, we describe the mechanics of the UBEM based on ST-GCN—a three-step process that aims to integrate physics-based and data-driven approaches, as shown in Figure 1. The objective of this framework is to improve UBEM by using ST-GCN that embeds the building interdependence as graph attributes. To construct and assess the feasibility of ST-GCN models, we start with analyzing the building interdependency and constructing a building interdependency graph network, followed by adding the graph features (building nodes and edges attributes) into the model. Through model comparison, this study aims to demonstrate that the hybrid model synthesizing physics-based knowledge and STGCN can improve data-driven UBEM performance in multiple aspects.

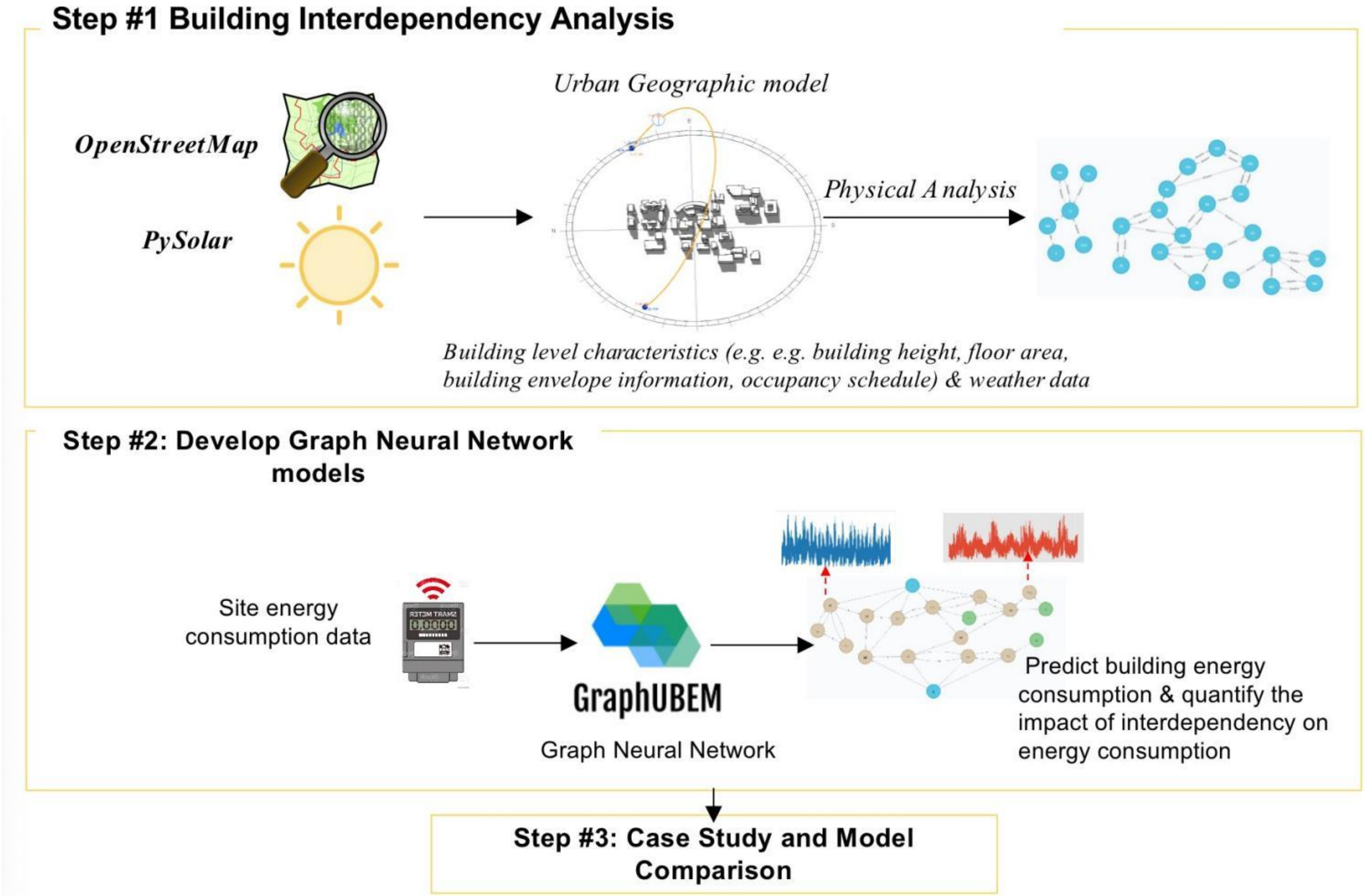

Figure 1. Methodology framework

*a. Building Interdependency Analysis*

The building interdependencies can be generated from multiple sources, which can be categorized as physics-based and occupant-based interdependency. Physics-based interdependencies typically include many types [4-7], for example, one building casts shadows on its nearby buildings, which will impact the energy consumption of its neighbors in different periods of a day; the long-wave radiation between building surfaces can also affect corresponding energy consumption; In addition, the relative location and building forms can impact the wind patterns and air humidity of micro-climate, which will also impact the energy consumption of these buildings [4, 5]. Many existing tools are available to conduct building solar analysis and wind simulation, for example, Revit, OpenFOAM, UrbaWind, etc. Occupant-based interdependencies are caused by human mobility among buildings. Regarding this, existing studies have discussed human-mobility-based geotagged social media data or mobile data [48]. Instead of directly considering all types of interdependencies, this paper focuses on the interdependency caused by building shadows and solar radiation and validates the impact of these interdependencies on the accuracy of the data-driven UBEM in forecasting. The procedure and framework developed in this paper can be used to analyze other building interdependencies in the future.

In this study, the solar-based interdependence is the key object, we use the date from OpenStreetMap[4] (OSM) and Pysolar[5] (as shown in Figure 1) to represent it. Firstly, we capture the polygon of target buildings from OSM Buildings, followed by searching the central node of each polygon (each of the nodes represents one target building). According to Erell et al. studies [49], if the distance between buildings is beyond three times of the building's height, the inter-building impact can be ignored. We adopted this assumption this paper to construct our initial graph. We connected the lines between different buildings, and an algorithm was developed to measure distances among buildings to filter the lines within effective range. Those filtered lines linking to nodes can be regarded as the graph edges, which are bidirectional. More specifically, the arrow from A to B means that Building A can influence Building B's energy consumption, see Figure 2. To describe the solar-based building interdependency (edge attributes), four parameters are obtained, namely solar radiation, the angle between building shadow and the connected line, distance, and binary variables. Real-time solar radiation is an important indicator that influences the heating and cooling demand of the individual building. When Building A casting a shadow on Building B, the solar radiation absorbed by B is inevitably less than before. Therefore, the energy consumption required by B is bound to be affected. To determine the scope of influence of each building, the 3 times of building's height is employed as the threshold. However, it is challenging and time-consuming to directly quantify dynamic impact of shadows. To solve this problem, we proposed to employ angle between building shadow and the connected line and distance in this study. It is obvious that solar-based impacts decline with the growth of the angle and distance. The binary variables are used to indicate that the shading of Building A is oriented to Building B (which is 1) or in a reverse direction (which is 0). The calculation of these parameters is shown below:

- *solar radiation: by running the Pysolar package, the time-series solar radiation can be captured based on GPS coordinates of buildings*
- *angle: the direction of shadow changes with sun azimuth, therefore, the time-series angle between building shadow and the connected line can be calculated based on time-series azimuth. In a physical system, the impact of building A on building B declines when the angle decreases. The angle quantifies the shadow impact to some extent (shown in Figure 2).*
- *binary: the binary value is used for qualitative analysis of shadow impacts. If the angle is acute, the binary value is 1; otherwise, it is 0.*
- *distance: the distance value can also be used to quantify the shadow impacts when B in A'sthreshold range.*

The four parameters are appended in a time-series array as the input of graph edge attributes. We use the four parameters to represent inter-building impacts, and then use data-driven methods to quantify the impact on building energy use.

[4] https://osmbuildings.org/
[5] https://pypi.org/project/pysolar/

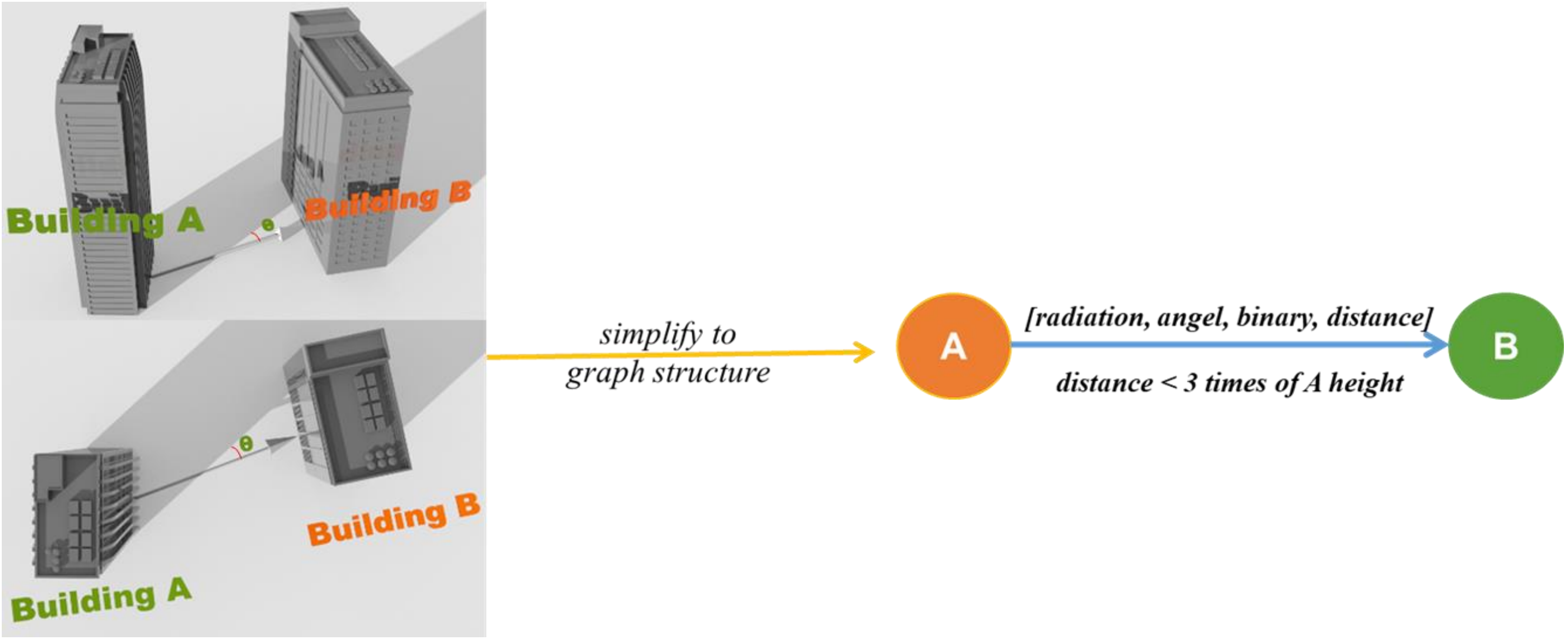

Figure 2. Solar-based building interdependency graph unit

Based on this analysis, we construct a directed weighted building dependency graph, and its basic unit is shown in Figure 2 (right). Each node represents a building, and the edges between the buildings represent the solar impacts. In addition to edge attributes, regular frequently-used building-level characteristics and weather features are also used as node attributes to predict building energy use.

*b. Constructing Graph Neural Network*

This paper aims to develop a data-driven model that can consider the interdependency between buildings. For each building, the number of its neighboring buildings is different, which causes complexity to transfer these interdependencies into fixed dimension attributes for model development. For example, a convolutional neural network (CNN) can only use on Euclidean data (e.g., image-2D grid and text-1D sequence). For each pixel or word, the number of connected objects is fixed, which enables its convolutional filters and pooling operators [41]. However, the building dependency graph is non-Euclidean with unfixed neighboring buildings. If using CNNs or other types of neural networks to operate the graph data and consider inter-building impacts, we need to consider all possible dependencies (n*n dependency matrix when the number of buildings is equal to n) and taking them as the features of each building, which will significantly increase features numbers, specifically when the number of buildings is large, and cause data sparsity. However, GNNs propagate information based on graph structure to reduce redundant computing and take the graph structure as part of features to update node labels with consideration of the status of its neighboring nodes [50]. The structure and information propagation mechanism comply with the need for building energy prediction in urban context. Specifically, to predict hourly building energy use, this paper adopts spatial-temporal graph convolutional network (ST-GCN). Spatial features reflect inter-building impacts and temporal features capture the time-dependency of energy use.

Since the graph structure reflects the solar-based building interdependency, utilization of this graph structure will be helpful to predict the energy consumption of the group of buildings. More specifically, we develop a novel framework which models the graph structure with Graph Neural Networks (GNN). Following ST-GCN, multiple ST-Conv blocks are applied to simultaneously extract the temporal features and spatial features of the buildings. The ST-Conv block contains two gated sequential convolution layers and one spatial graph convolution layers between them. The gated sequential convolution layer apply gated linear units (GLU) as activation function. And the spatial graph convolution layers are based on graph convolutional network. Let $\mathbf{X}_{\mathrm{t}}$ represents the features of all the buildings at time point $t$**,** the graph convolution can be formulated as:

$$\mathbf{H_t} = \sigma\left(\widetilde{\mathbf{D}}^{-\frac{1}{2}}\widetilde{\mathbf{A}}\widetilde{\mathbf{D}}^{-\frac{1}{2}}\mathbf{X}_{\mathrm{t}}\mathbf{W}\right),$$

Where $\widetilde{\mathbf{A}} = \mathbf{A} + \mathbf{I}$ and $\mathbf{A}$ is the adjacency matrix of the graph structure. $\widetilde{\mathbf{D}}$ is the degree matrix of $\widetilde{\mathbf{A}}$**.** $\mathbf{W}$ represents the learnable parameters of the graph convolution layer. $\sigma$ is the activation function, which is ReLU in implementation. The adjacency matrix $\mathbf{A}$ denotes the structure of the graph, and $A_{ij}$ denotes the relation between node $v_i$ and $v_j$. A larger $A_{ij}$ indicates tighter relation between them. As the relation between the buildings is solar-based, it is promising to adopt a time dependent adjacency matrix to reflect the change of relations along time. Therefore, different from ST-GCN, we propose to learn the adjacency matrix $A$ at time $t$ from the features of the buildings $\mathbf{X}_{\mathrm{t}}$. The process can be formally written as:

$$A_{ij} = \begin{cases} 0, & if\ v_i\ and\ v_j\ are\ not\ linked \\ MLP\left([x_i^t, x_j^t]\right), & else' \end{cases}$$

Where $x_i^t$ and $x_j^t$ are the features of building $v_i$ and $v_j$ at time point $t$. Multi-Layer Perception (MLP) is applied to learn the weight of the existing edges in the graph.

*c. Validation and Model Comparison*

This paper uses Root Mean Square Error (RMSE) and Mean Absolute Percentage Error (MAPE) to assess the proposed model accuracy and compare it with other common data-driven energy prediction models to test the significance of involving building interdependency into a data-driven model and associated model performance improvement. RMSE and MAPE was show in Formula 1 and Formula 2, respectively:

$$RMSE = \sqrt{\sum_{i=1}^{n} \frac{(y_i - \widehat{y_i})^2}{n}} \tag{1}$$

$$MAPE = \left(\frac{1}{n}\sum_{i=1}^{n} \frac{|y_i - \widehat{y_i}|}{\widehat{y_i}}\right) * 100\% \tag{2}$$

Where $y_i$ is the model predicted energy consumption value, $\widehat{y_i}$ is the actual value, *n* is the number of measurements. To compare model performance, we compare it with six frequent-used and proven effective algorithms for time series forecasting problem. last-hour model, average model, linear-regression model [51], Multilayer perceptron network (MLP) model [52], eXtreme Gradient Boosting (XGBoost) model [53], and Gated Recurrent Unit (GRU) network model [54]. The introduction and construction of these models is summarized in Table 1.

Table 1: Summary of compared data-driven models

| Model name | Description |
|---|---|
| Last hour | Estimates the future energy consumption using that in the hour prior to it |
| Average | Estimates the future energy consumption by the average value of that in the last 12 hours |
| Linear regression | Builds one regressor for each building to predict its energy use, which is based on characteristics of buildings and the weather data. The regressor is constructed as a linear model. |
| MLP | MLP is a class of feed forward artificial neural network model. Each block of it is composed of one linear layer and one non-linear activation layer, and a MLP is constructed by stacking multiple blocks. In this project, for each building, we construct a MLP containing two such blocks, and use last hour's features as inputs to predict the energy use for the following hour. |
| XGBoost | XGBoost represents the gradient boosting ensemble algorithm, which is a widely used ensemble model. An ensemble model is composed of multiple base models, and each base model is expected to capture one aspect of the knowledge. Hence utilizing those base models al-together is expected to be more accurate in predictions. Base model is usually implemented as a smaller model, and decision tree is adopted in XGBoost algorithm. We used it to predict short-term building energy use, and iteratively add new base learners to reduce the error between predictive values and labeled data. XGBoost is built for each building and output timestamp. |
| GRU | GRU is one of the most commonly used type of Recurrent Neural Network (RNN) [55], which is developed to solve the gradient vanishing and gradient explosion problems in standard RNN networks. It extends classical RNN through adding an update gate and a reset gate inside each block, hence is better at capturing long-term dependencies. Many studies have shown the effectiveness of GRU for time-series forecasting [54]. In this approach, for each building we use a GRU to aggregate its historical features, before passing it to a linear layer for prediction. |

*d. Case Study*

To test the performance of spatial-temporary graph convolutional networks and validate the importance of integrating building interdependency as input for data-driven models, we run and compare the energy prediction of various models for building energy consumption using the central area of an urban university campus located in the downtown area of Atlanta, Georgia. This study area contains 26 buildings, including classrooms, offices, and libraries. As a proof of concept, we simplify inter-building dependencies and focus on solar-based dependency in this paper. OpenStreetMap and PySolar are used to construct urban models and perform solar analysis. After calculations, 42 dependency edges were identified between buildings. The building interdependency graph of the test project is shown in Figure 3. Each node represents a building, and directed edges represent their solar impacts with width of the edges representing the impact level.

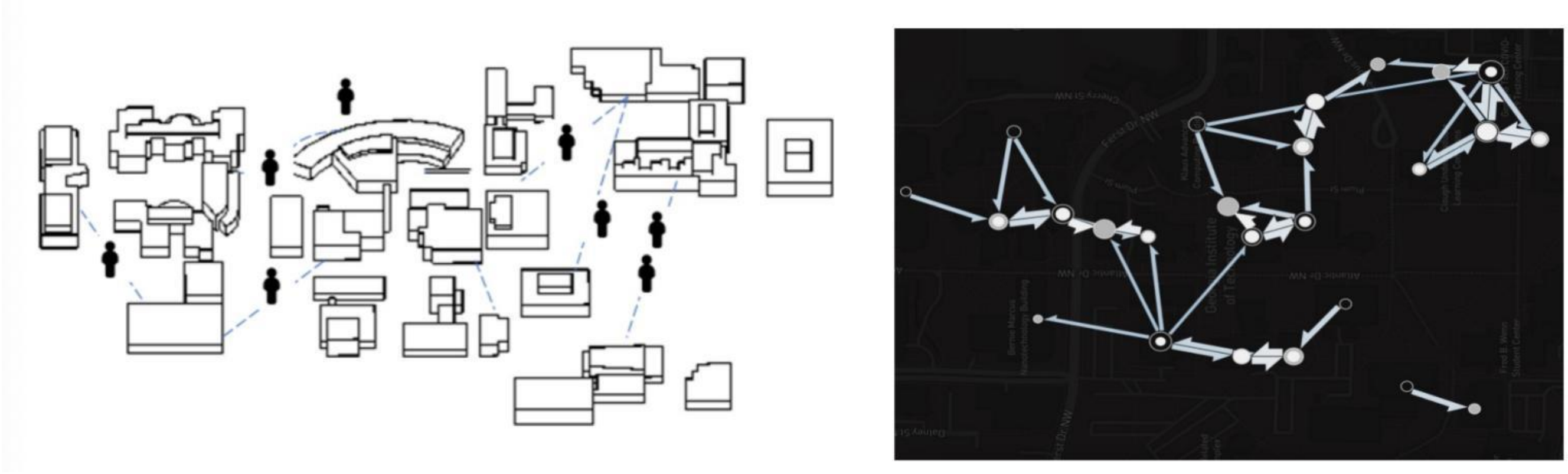

a. the area of case study　　　　b. representation of graph network

Figure 3. building graph network of case study[6]

The input attributes of UBEM incorporate a number of aspects covering the building interdependency, physical characteristics, weather conditions, and historical building energy consumption data, as shown in Table 2 (building interdependency has shown in Figure 3). Firstly, the building graph structure is constructed as shown Figure 3, followed by adding the building physical characteristics to each node, which are chosen according to the important parameters mentioned in past literature for data-driven building energy modeling. The information of building size is encoded in the parameters ‘Gross Floor Area’, ‘Number of Floors’, ‘Internal space area’, which indicate the level of building usage (e.g., how much space is air-conditioned, how many occupants possibly in the building). The building design and construction characteristics, represented by thermal mass, building envelope area, window-wall ratio, are also embedded since these are highly correlated to heat transfer between indoor and outdoor environments. Finally, we incorporate the LEED certification rating and the occupancy year as strong indicators and adjustments of the energy efficiency of buildings (e.g., building insulation level, building system coefficient of performance, etc. Occupancy year is processed as a categorical variable, for which

[6] The visualized graph network can be found in:
https://flowmap.blue/1RtHXxI_HrV4V2mmd6kIyJNXjVkF9A20S27BBYAr5q0Q
Note: the size of visualized edge weight increases as the distance decreases.

22 different years are divided into six periods based on the energy code change timeline in Georgia. School calendar is also collected to identify occupant use patterns, which are divided into five categories (class day, weekend, in semester holiday (e.g., spring break), summer semester, and holiday (winter and summer breack) to represent use pattern difference. Thermal mass is classified into three categories: light, medium, and high, based on envelope materials, which depend on the portion of glazing in the building enclosure and construction materials (e.g., a building with a complete curtain wall is classified as light thermal mass but that with mostly concrete will be classified as a high thermal mass building). Thirdly, the weather conditions were shown to impact the building energy consumption in a continuous manner. In this study, we utilized the recorded hourly weather data from the local campus weather stations, including the air temperature, relative humidity, wind speed, and solar density, all of which are significant parameters that affect building heat transfer with outdoor environment and HVAC energy consumption. Since the building energy consumption is demonstrated to be significant time dependency, the lagged energy consumption ahead of the prediction time window of each building was also employed to forecast corresponding time-series energy consumption, respectively.

Table 2. Data frame used for the data-driven UBEM

| Features | Definition | Data source | Data type |
|---|---|---|---|
| Building Interdependency | Inter-building solar impact | PySolar and OpenStreetMap | Graph |
| Building & Occupant characteristics | Occupancy year | Client (University facility management department) | categorical |
| | LEED certification | Client | binary |
| | Gross floor area | Client | numerical |
| | Number of floors | Client | numerical |
| | Window-wall ratio | Google Map | numerical |
| | Building envelope area | DynaMap and Revit | numerical |
| | Thermal mass level (light, medium, high) | Google Map | categorical |
| | Internal space area | Client | numerical |
| | Building occupancy pattern | Public school calendar | Categorical (three types: class day, weekends, and holiday) |
| Weather Data | Hourly air temperature, relative humidity, wind speed, and solar density | Public weather station | Time-series numerical |
| Building energy consumption | Hourly energy consumption of each building | Client | Time-series numerical |

## RESULT AND ANALYSIS

This section introduces the predicted energy consumption results and model comparison. After running the different data-driven algorithms, we calculate the Root Mean Square Error (RMSE) and Mean Absolute Percentage Error (MAPE) per Formula discussed in Section 4. Due to the abnormal fluctuation of original data (the two building were in renovation in 2016&2017), we deleted two buildings results in the graph network (Building ID 10001 and 10002 in graph), 24 buildings were left finally.

### *a. Model Comparison*

To evaluate the accuracy of the data-driven UBEMs, we use the overall RMSE and MAPE for 24 buildings as indicators as shown in Table 3. It is obvious that overall ST-GCN outperforms other machine learning models. Through comparing the statistical performance among those models, it is found that the prediction of future energy consumption is not trivial that can be easily recovered from historical time-series energy use information. This reflects the performance deficiency of only taking the last hour and the average of previous consumption as inputs of heuristic methods compared to than other data-driven approaches (MLP, XGBoost, GRU and ST-GCN).

Table 3. RMSE and MAPE of selected models

| Model | RMSE | MAPE |
|---|---|---|
| Last hour model | 43.81 | 16.67% |
| Average model | 50.32 | 14.04% |
| Linear regression | 34.49 | 30.47% |
| Multi-layer perceptron (MLP) | 25.20 | 7.85% |
| XGBoost | 22.52 | 6.93% |
| GRU | 28.58 | 7.72% |
| ST-GCN | 18.56 | 5.21% |

To further demonstrate the performance of ST-GCN compared to other methods, we listed the RMSE of 24 buildings in each model (see Figure 4). A similar phenomenon can also be observed in the RMSE of individual buildings. Generally, ST-GCN outperforms other models and GRU, XGBoost, MLP, and ST-GCN are more reliable than other models. The variance of RMSE of these four models is obtained to reflect the model robustness, GRU: 158, XGBoost: 168, MLP: 101, ST-GCN: 92; this further indicates that the ST-GCN achieves the least variance, therefore, being more robust in building energy usage modeling.

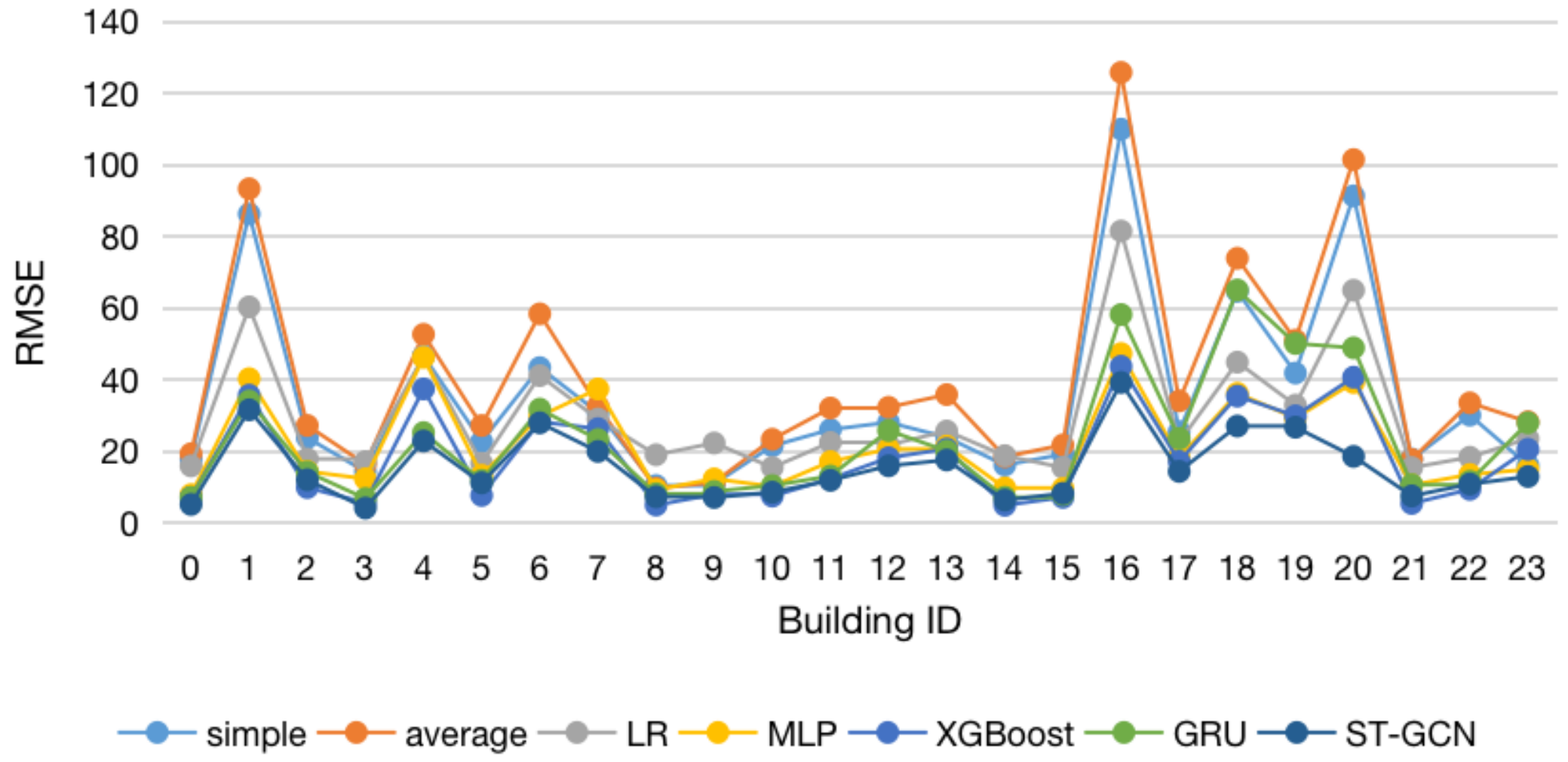

Figure 4. The RMSE of 24 buildings for different data-driven models

Meanwhile, MAPE is also employed to evaluate the accuracy in energy simulation models. To validate the effectiveness of data-driven models, the MAPE of the best four models was calculated as presented in Figure 5. In order to contextualize the baseline results, we compare them to the ASHRAE Standard Guideline 14 [56], which defines metrics and guidelines for hourly building energy simulation for individual buildings. Based on the ASHRAE standard, the acceptable ranges for an energy simulation require an accuracy of 30% for hourly intervals. It is obvious that the four models mostly meet the requirements of ASHRAE, and the distribution pattern of accuracy is similar to the RMSE of buildings. ST-GCN is the most outstanding model in hourly energy prediction (shown in Figure 5), and most of the intervals even fall within 5%. These results fully illustrate the ST-GCN is highly accurate in high-resolution modeling of building energy consumption.

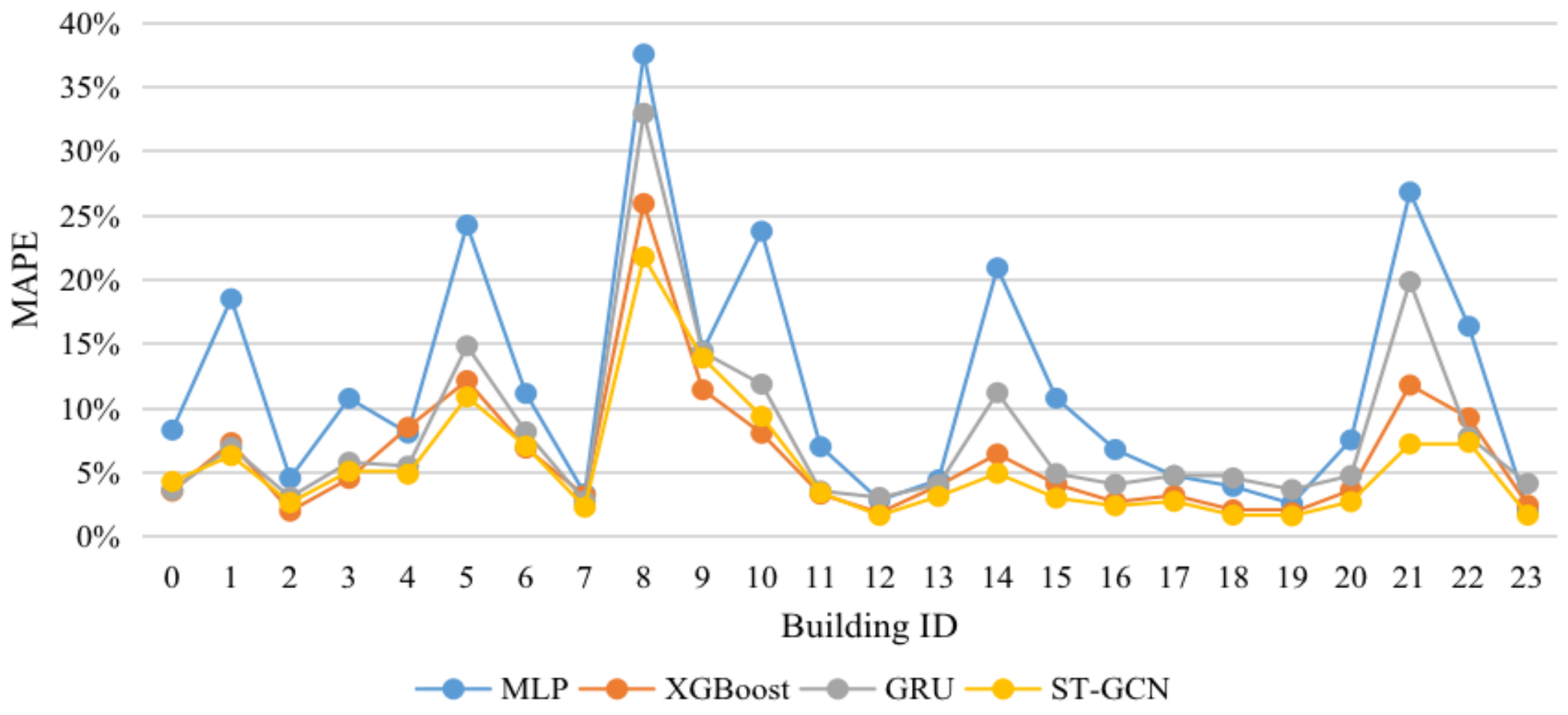

Figure 5. The MAPE of 24 buildings for three models

b. *Model Interpretation*

To further indicate the interpretability of ST-GCN, we generated the distribution of seasonal RMSE of 24 buildings for evaluation (see Figure 6). Similarly, the GRU, MLP, XGBoost and ST-GCN outperform overall, and the ST-GCN is still the best compared to all others. When discussed seasonal difference, heating and cooling load are typically high in summer and winter compared with spring and fall seasons. Since heating and cooling load are sensitive to solar radiation in summer and winter [56], the shading impact of nearby buildings on building energy use will be more significant. For conventional data-driven models, they are not able to capture solar-based interdependency variation, while ST-GCN considers the building interdependency. Therefore, we assume the predicted results of ST-GCN in summer and winter can be further optimized than in winter and spring. To validate the hypothesis, we calculate the improvement rate of the RMSE of the ST-GCN model with other models in four seasons, as shown in Table 4. The result supports our hypothesis. When integrating solar-based interdependency into edge attributes, the winter and summer RMSE of ST-GCN in most of the buildings have significant improvement when simulating hourly energy consumption, whereas RMSE in spring and fall change relatively small (102% in summer, 98% in winter, 71% in spring, 59% in fall).Considering the solar-based interdependency in the data-driven models is more necessary and explains why ST-GCN improves more in winter and summer.

Table 4. The improved rate of RMSE of ST-GCN vs. each machine learning model

| | Last Hour | Average | LR | MLP | XGBoost | GRU | **Average improvement** |
|---|---|---|---|---|---|---|---|
| Spring | 118% | 155% | 81% | 33% | 7% | 32% | **71%** |
| Summer | 176% | 179% | 106% | 59% | 21% | 69% | **102%** |
| Fall | 83% | 131% | 67% | 25% | 14% | 31% | **59%** |
| Winter | 143% | 187% | 115% | 36% | 27% | 77% | **98%** |

| Building ID | 0 | 1 | 2 | 3 | 4 | 5 | 6 | 7 | 8 | 9 | 10 | 11 | 12 | 13 | 14 | 15 | 16 | 17 | 18 | 19 | 20 | 21 | 22 | 23 | |
|---|---|---|---|---|---|---|---|---|---|---|---|---|---|---|---|---|---|---|---|---|---|---|---|---|---|
| spring | 18.58 | 78.12 | 24.99 | 14.77 | 53.44 | 22.01 | 50.69 | 31.94 | 8.32 | 6.44 | 22.68 | 26.64 | 25.48 | 24.36 | 15.90 | 19.50 | 114.13 | 28.13 | 67.12 | 37.20 | 75.30 | 17.10 | 32.89 | 17.96 | Last hour |
| summer | 20.93 | 91.78 | 23.70 | 15.81 | 54.16 | 23.76 | 20.10 | 47.94 | 11.22 | 7.39 | 30.37 | 29.72 | 35.76 | 20.97 | 19.50 | 21.07 | 115.52 | 26.87 | 85.00 | 62.98 | 126.45 | 17.69 | 27.37 | 14.17 | |
| fall | 14.74 | 96.85 | 22.60 | 13.43 | 36.93 | 23.27 | 46.28 | 20.32 | 9.75 | 10.35 | 17.56 | 24.86 | 24.32 | 25.56 | 14.89 | 17.40 | 102.76 | 21.96 | 51.13 | 37.66 | 84.85 | 18.91 | 28.09 | 15.78 | |
| winter | 19.09 | 73.43 | 24.38 | 14.22 | 41.72 | 22.75 | 38.93 | 24.10 | 14.63 | 19.37 | 15.45 | 24.00 | 31.80 | 23.04 | 15.97 | 20.49 | 110.48 | 25.86 | 61.12 | 31.63 | 91.87 | 17.85 | 31.36 | 15.46 | |
| spring | 20.68 | 90.52 | 28.30 | 16.90 | 58.56 | 27.18 | 61.47 | 32.89 | 7.49 | 7.23 | 25.64 | 32.84 | 29.51 | 33.79 | 18.25 | 22.71 | 137.24 | 36.89 | 78.69 | 48.26 | 86.92 | 16.23 | 35.50 | 21.83 | Average |
| summer | 19.34 | 90.62 | 24.99 | 14.48 | 51.44 | 25.02 | 30.80 | 44.93 | 9.70 | 8.05 | 28.96 | 32.59 | 35.38 | 22.22 | 19.28 | 19.46 | 120.28 | 28.78 | 78.79 | 59.77 | 120.11 | 16.13 | 27.39 | 30.09 | |
| fall | 17.68 | 101.02 | 28.26 | 17.16 | 50.82 | 28.96 | 69.39 | 24.90 | 9.23 | 11.69 | 20.84 | 32.79 | 29.88 | 44.99 | 18.59 | 21.77 | 120.01 | 34.98 | 63.42 | 49.85 | 101.11 | 19.53 | 34.62 | 34.11 | |
| winter | 20.18 | 85.44 | 25.56 | 16.17 | 45.14 | 26.20 | 48.84 | 31.18 | 13.88 | 18.49 | 16.70 | 29.02 | 38.95 | 30.31 | 18.47 | 22.66 | 118.74 | 31.82 | 78.83 | 49.74 | 108.69 | 18.92 | 33.53 | 25.08 | |
| spring | 15.45 | 52.26 | 18.15 | 18.03 | 47.75 | 16.88 | 47.71 | 30.65 | 18.82 | 22.06 | 13.73 | 20.95 | 19.13 | 25.88 | 18.79 | 14.62 | 80.15 | 27.07 | 44.17 | 26.19 | 55.31 | 14.44 | 19.24 | 23.52 | LR |
| summer | 13.92 | 60.07 | 13.72 | 14.08 | 33.95 | 15.26 | 23.21 | 30.98 | 17.54 | 20.89 | 15.57 | 17.79 | 20.28 | 24.95 | 14.70 | 12.08 | 88.17 | 16.79 | 61.57 | 52.11 | 89.64 | 14.51 | 11.40 | 23.91 | |
| fall | 16.53 | 71.24 | 18.52 | 17.68 | 48.94 | 16.58 | 43.22 | 29.39 | 19.02 | 22.34 | 15.83 | 24.32 | 21.74 | 25.05 | 20.98 | 16.86 | 79.01 | 20.38 | 36.44 | 30.57 | 59.23 | 15.42 | 19.33 | 24.62 | |
| winter | 17.95 | 51.85 | 20.10 | 17.56 | 50.34 | 18.28 | 37.12 | 22.78 | 21.37 | 24.62 | 18.33 | 26.16 | 27.94 | 27.51 | 18.21 | 17.57 | 82.97 | 22.51 | 42.44 | 24.10 | 65.02 | 18.56 | 20.53 | 21.66 | |
| spring | 7.35 | 26.66 | 15.37 | 12.62 | 47.85 | 13.29 | 33.66 | 37.57 | 8.33 | 10.84 | 7.89 | 16.43 | 17.71 | 19.49 | 10.23 | 8.31 | 46.06 | 21.68 | 30.34 | 25.53 | 51.57 | 9.54 | 15.17 | 15.56 | MLP |
| summer | 7.87 | 31.41 | 18.71 | 17.48 | 48.77 | 17.82 | 23.23 | 46.45 | 10.93 | 15.29 | 12.07 | 17.65 | 22.17 | 29.03 | 13.24 | 9.31 | 39.87 | 14.21 | 39.51 | 35.88 | 38.41 | 13.59 | 12.07 | 12.54 | |
| fall | 8.94 | 58.43 | 12.23 | 10.44 | 49.95 | 11.66 | 31.19 | 37.64 | 7.90 | 10.27 | 12.05 | 18.75 | 21.61 | 22.00 | 8.59 | 9.38 | 50.01 | 18.96 | 32.41 | 32.65 | 28.98 | 10.61 | 13.78 | 15.56 | |
| winter | 7.41 | 23.30 | 11.82 | 9.37 | 30.57 | 11.62 | 24.94 | 25.14 | 12.12 | 15.64 | 8.81 | 15.12 | 22.98 | 14.66 | 7.42 | 12.90 | 51.47 | 12.10 | 48.95 | 21.78 | 28.21 | 9.75 | 11.40 | 15.40 | |
| spring | 6.85 | 23.15 | 10.20 | 6.05 | 35.08 | 5.88 | 32.08 | 25.67 | 3.77 | 6.08 | 5.90 | 10.95 | 15.38 | 20.77 | 5.14 | 7.40 | 43.21 | 21.36 | 26.94 | 24.60 | 44.28 | 3.11 | 9.23 | 16.70 | XGBoost |
| summer | 6.93 | 36.08 | 9.98 | 4.32 | 30.53 | 8.79 | 18.00 | 30.34 | 5.15 | 6.03 | 7.65 | 12.16 | 16.34 | 20.94 | 3.94 | 6.16 | 37.20 | 12.17 | 44.73 | 32.99 | 39.51 | 2.55 | 7.52 | 17.09 | |
| fall | 6.85 | 49.60 | 10.84 | 6.28 | 41.32 | 8.33 | 29.58 | 28.81 | 5.31 | 7.52 | 9.91 | 13.67 | 17.23 | 22.36 | 5.79 | 7.19 | 48.44 | 17.13 | 31.75 | 33.53 | 38.77 | 7.66 | 9.92 | 27.87 | |
| winter | 8.50 | 19.75 | 7.39 | 5.28 | 40.50 | 8.84 | 26.19 | 16.84 | 6.66 | 13.00 | 4.66 | 12.80 | 25.99 | 17.07 | 4.07 | 7.98 | 41.51 | 12.05 | 45.83 | 29.27 | 38.67 | 6.47 | 10.73 | 12.66 | |
| spring | 7.33 | 22.20 | 13.25 | 7.36 | 23.61 | 11.83 | 32.49 | 22.63 | 6.62 | 5.97 | 8.05 | 12.03 | 23.13 | 20.21 | 7.07 | 8.47 | 47.17 | 26.91 | 58.73 | 44.30 | 47.26 | 9.50 | 13.08 | 25.33 | GRU |
| summer | 6.87 | 28.69 | 10.08 | 7.44 | 27.22 | 12.81 | 26.46 | 30.85 | 8.34 | 6.47 | 12.57 | 16.93 | 27.45 | 17.38 | 7.77 | 8.43 | 65.53 | 21.02 | 72.27 | 62.73 | 48.18 | 10.69 | 10.30 | 32.71 | |
| fall | 6.66 | 45.43 | 14.62 | 6.58 | 25.59 | 11.84 | 32.73 | 19.97 | 7.73 | 7.91 | 12.95 | 11.97 | 24.16 | 19.56 | 7.04 | 7.19 | 64.43 | 19.92 | 64.23 | 49.52 | 43.79 | 12.00 | 11.00 | 30.33 | |
| winter | 8.20 | 32.61 | 18.76 | 6.10 | 26.19 | 11.74 | 33.76 | 22.23 | 11.45 | 15.44 | 6.52 | 13.61 | 32.19 | 23.78 | 7.55 | 8.32 | 57.70 | 26.33 | 71.40 | 48.56 | 61.44 | 10.59 | 11.70 | 23.01 | |
| spring | 7.96 | 19.72 | 12.12 | 6.27 | 19.04 | 10.83 | 30.90 | 21.35 | 5.19 | 5.17 | 6.21 | 11.01 | 13.56 | 18.43 | 6.35 | 8.58 | 37.10 | 17.92 | 27.16 | 24.06 | 39.71 | 5.74 | 12.39 | 15.15 | ST-GCN |
| summer | 9.95 | 22.67 | 8.06 | 4.61 | 24.72 | 7.47 | 15.81 | 28.33 | 7.96 | 7.05 | 8.45 | 12.84 | 14.82 | 23.10 | 5.63 | 8.35 | 28.25 | 10.05 | 27.01 | 31.44 | 13.80 | 5.69 | 7.80 | 9.86 | |
| fall | 7.79 | 44.96 | 14.11 | 7.52 | 27.29 | 11.75 | 31.69 | 16.22 | 7.24 | 7.62 | 11.71 | 12.99 | 14.82 | 16.04 | 7.56 | 8.44 | 45.36 | 14.14 | 27.97 | 30.89 | 26.94 | 8.23 | 11.48 | 13.54 | |
| winter | 7.94 | 26.16 | 11.42 | 4.75 | 18.60 | 8.05 | 23.49 | 13.17 | 10.33 | 13.62 | 5.48 | 10.86 | 22.81 | 12.10 | 4.98 | 7.85 | 30.11 | 10.85 | 25.73 | 16.34 | 17.46 | 10.84 | 9.40 | 10.10 | |

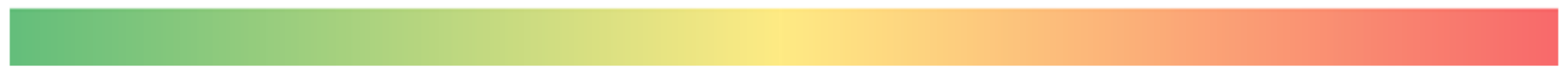

Figure 6. Distribution of seasonal RMSE of 24 Buildings

Also, we extracted the indegrees of the graph (as shown in Figure 8). The indegree represents the number of adjacent buildings that potentially affect the target building energy consumption. Hence, a larger indegree of a building indicates more significant impacts of these interdependencies on energy use of the target buildings in energy prediction. As only ST-GCN accounts for the inter-building impacts in energy prediction, we expect to observe more significantly improved performance of ST-GCN compared to other approaches in energy prediction of buildings with larger indegrees. To test this hypothesis, we compare the performance of the other two best models (GRU and XGBoost) with ST-GCN and buildings' indegree. By comparing with the distribution pattern of improved MAPE on different indegrees, it is obvious that the accuracy improvement increases significantly with indegrees (see Table 5). Both the average improvement of MAPE for GRU and XGBoost reach peaks while indegree is three. Furthermore, we explored the relationship between MAPE improvement individual buildings. Generally, the improved MAPE of these buildings with the highest indegrees are generally outstanding among other models. For example, Building 3, 5, 8, 10 have a significant improvement on MAPE; whereas the improvement of buildings without indegrees are relatively small (see Building 9, 12, 13, 14). One phenomenon worth noting is the great accuracy enhancement on Building 21, with only 2 indegrees. By exploring the physical relationship of buildings (as shown in Figure 10(a)), we found two surrounding buildings close to Building 21, (one located on the west of Building 21 and the other located on the south), which can have a huge impact. Even though the XGBoost and GRU in most buildings are highly accurate, the ST-GCN consistently improves the predicted results. The exceptions are Building 4 and 9, which have one or no indegree. For Building 4, the adjacent building located on the northwest casts on a very small shadow on building 4 (see Figure 10(b)).

Table 5. The improved rate of MAPE of ST-GCN vs. GRU and XGBoost in different indegrees

| Indegree | Node count | GRU | XGBoost |
|---|---|---|---|
| 0 | 6 | 4% | 1% |
| 1 | 4 | 7% | 2% |
| 2 | 6 | 10% | 3% |
| 3 | 6 | 11% | 5% |

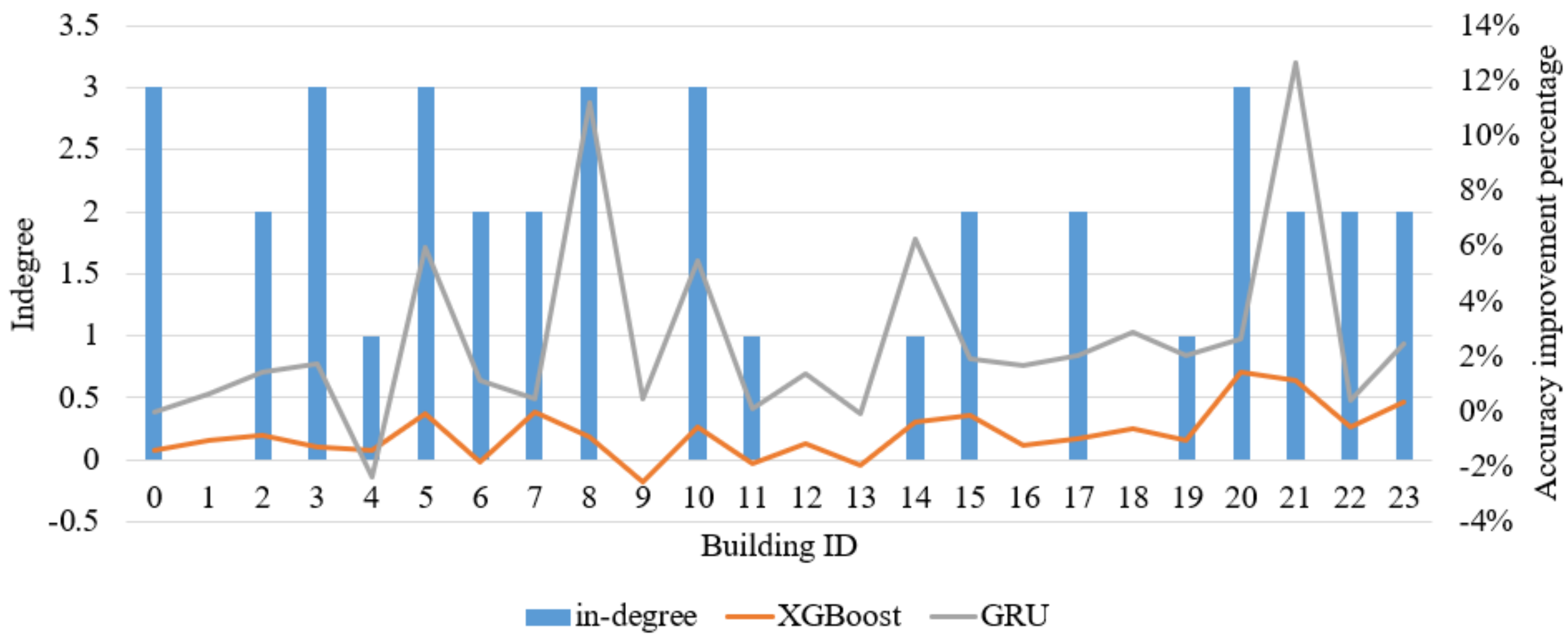

Figure 9. The indegree of the graph and improvement percentage

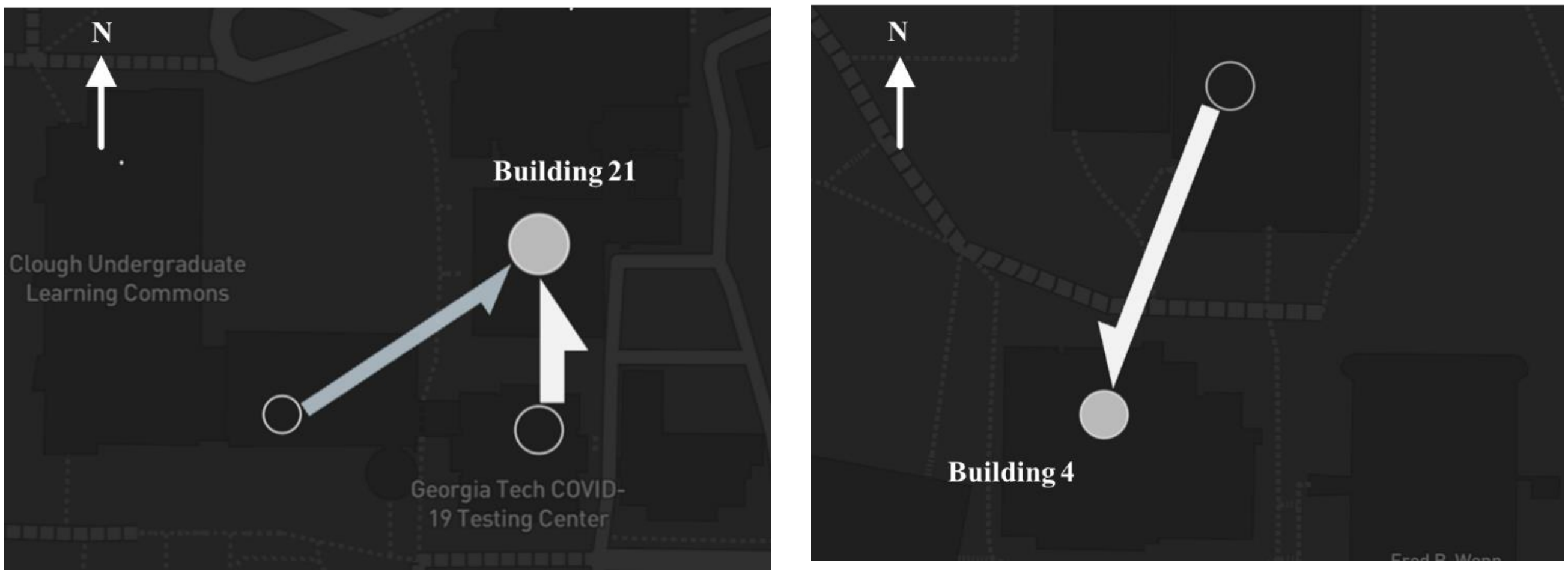

(a) Building 21 (b) Building 4

Figure 10. The physical graph of Building 21 and 4

## DISCUSSION AND LIMITATION

Capturing complex building interdependency in the urban context is challenging but significant to enhance the accuracy of both physics-based and data-driven models in building energy modeling. This study aims to take a first step to overcome this challenge in urban energy modeling and uncover the advantages of data-driven UBEM: a time-series urban building energy model integrated with the spatio-temporal graph convolutional network algorithm and the solar-based interdependency. We utilized buildings in a university campus as the case study to evaluate the feasibility of the proposed ST-GCN model in energy modeling. The results indicate that the developed ST-GCN model outperforms both heuristic and other machine learning models. The performance of the ST-GCN model also exceeds the benchmark accuracy indicated in the ASHARE Standard. Compared to other purely data-driven models, the ST-GCN model demonstrates explicitly improved performance, especially in summer and winter season when building heating and cooling load is high. By comparing the ST-GCN performance and graph attributes (e.g., indegree), we found that the performance improvement of ST-GCN model is positively correlated with the indegree of the node, which highlights the necessity of involving building interdependency in urban building energy prediction. Previous studies have discussed building energy use can be affected by a wide range of factors, such as microclimates and inter-building impacts. By integrating the engineering knowledge into data-driven models, the accuracy and explanation of data-driven UBEM will be significantly improved, especially in a complex urban environment.

Nevertheless, there are still some limitations in this study. Firstly, we evaluated these data-driven models through 26 co-located campus buildings for energy consumption in 2016 and 2017, so the evaluated time spans and building types are relatively limited. Future work is required to further validate the modeling approach on other case studies with more diverse building stocks and urban morphologies to ensure reliability of the proposed model. Furthermore, our model was only applied in the urban (campus) scale, its effectiveness of energy prediction in a geographic area of larger scale (e.g., central business districts, suburban areas) need to be evaluated. Moreover, various types of building interdependency exist in urban environments, although only the solar-based interdependency is included in this study. As such, acknowledge that our case study is not fully representative of a large real city, but nonetheless the results provide a high-level validation of how our proposed model would perform on a small and dense downtown area that is common in many parts of the east America. More efforts need to be made in the future to include multiple aspects of building interdependency and microclimates to represent the real urban environment. At last, a heterogeneous graph neural network containing various categories of buildings, building interdependencies and microclimates is expected to be constructed which can be applied on different cities.